# Going Wider: Recurrent Neural Network With Parallel Cells

Danhao Zhu, Si Shen, Xin-Yu Dai and Jiajun Chen
Department of Computer Science, Nanjing University, China

Abstract: Recurrent Neural Network (RNN) has been widely applied for sequence modeling. In RNN, the hidden states at current step are full connected to those at previous step, thus the influence from less related features at previous step may potentially decrease model's learning ability. We propose a simple technique called parallel cells (PCs) to enhance the learning ability of Recurrent Neural Network (RNN). In each layer, we run multiple small RNN cells rather than one single large cell. In this paper, we evaluate PCs on 2 tasks. On language modeling task on PTB (Penn Tree Bank), our model outperforms state of art models by decreasing perplexity from 78.6 to 75.3. On Chinese-English translation task, our model increases BLEU score for 0.39 points than baseline model.

**1.Introduction**

Recurrent neural network (RNN) has been one of the most powerful sequence models in natural language processing. Many important applications achieve state of the art performance with RNN, including language modeling(Mikolov et al.,2010; Zaremba et al., 2014 ), machine translation (Luong and Manning, 2016; Wu et al., 2016) and so on.

The features learned by RNN are stored in the hidden states. At each time step, the cell extracts features from data and updates its hidden states. The left side of figure 1 concisely shows the transition of hidden states in naïve RNN. We can see that all units at the previous step are fully connected to all units at the current step. Thus, each pair of features can affect each other. Such design is not reality because many features are not that related. The influence between unrelated features may harm the learning ability of models. We can expect learning models automatically set the weight of all unnecessary connections to zero. However, in practice, because the data size is limited and the algorithms are not that strong, these unrelated connections will harm the learning ability. For example, that is why we have to do feature selections before training.

To address the problem, many successful neural models benefit from replacing global connection with local connection. For example, Long Short-Term Memory(LSTM) (Hochreiter and Schmidhuber, 1997) and Gated Recurrent Units(GRU) (Cho et al., 2014) have been the most popular RNN cells. In the core, such models use gates to control the data flow, allow part of connections to be activated. Another example is Convolution Neural Networks(CNN)(Lecun et al, 1998), one of the most successful models in deep learning nowadays. CNN uses local receptive fields to extract features from previous feature map. With local receptive fields, neurons can extract elementary visual features such as oriented edges, end-points, corners.

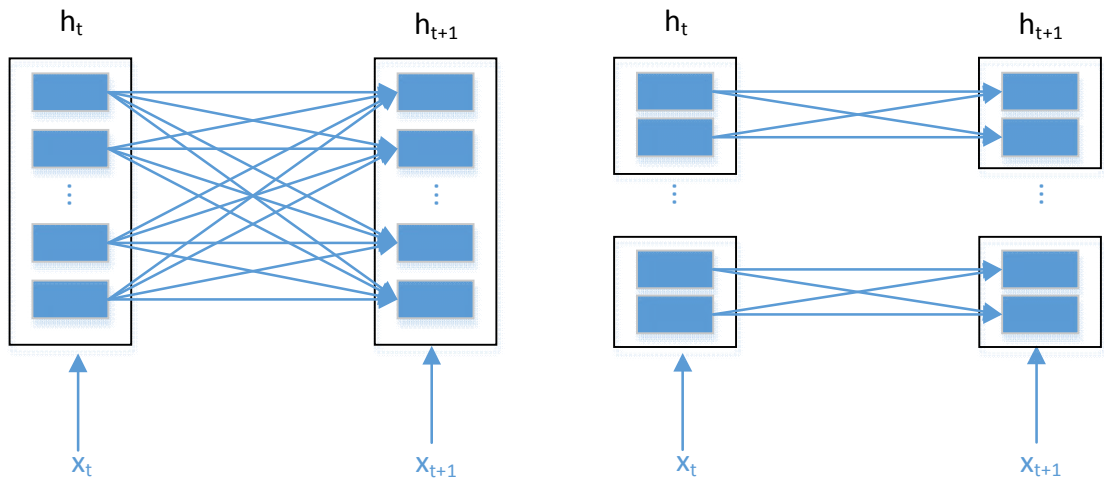

Figure 1: The Transition of RNN hidden states. The left side is naïve RNN. The right side is RNN with Parallel Cells. $h_t$ and $x_t$ denote hidden states and input vector at time step t respectively.

In this line, we propose a model named RNN-PCs (Recurrent Neural Network with Parallel Cells) to improve the learning ability of RNN, meanwhile to largely reduce the number of parameters. The model replaces global recurrent connections with small local connections. We replace a single large cell with many smaller ones (the hidden states have less units). As in the right part of figure 1, the hidden states are no longer full connected but only connected in a local manner. Figure 2 shows an unrolled 2 layer RNN with 2 parallel cells. In each layer of RNN, each small cell extracts and saves features from outputs of previous layer independently. And the outputs of small cells are concatenated as the output of current layer. The design has several advantages. First, each cell extracts features independently. Features in one cell transferred by its own recurrent connection and are not impacted by features in other cells. Second, parameters for recurrent connection decrease significantly. Also, one cell has to be placed entirely in one GPU. Now we have multiple small cells rather than one single larger cell, thus we can place these cells in different GPUs to optimize the training. Thirdly, Parallel Cells(PCs) does not modify the inner structure of cells, as a result, PCs can be used along with any type of RNN cells, such as LSTM.

Indeed, our idea of RNN-PCs is inspired by the multi-filter mechanisms in CNN. CNN uses multi filters to generate multi feature maps. Next, the feature maps are stacked as the input of following steps. RNN-PCs works in a similar manner. We use multi cells to extract different features from current inputs. And the outputs are stacked as the inputs of following steps. On the empirical stage, CNN shows that different filters are running for different features. We also make an empirical study on language modeling task, and show that different cells offer different set of features.

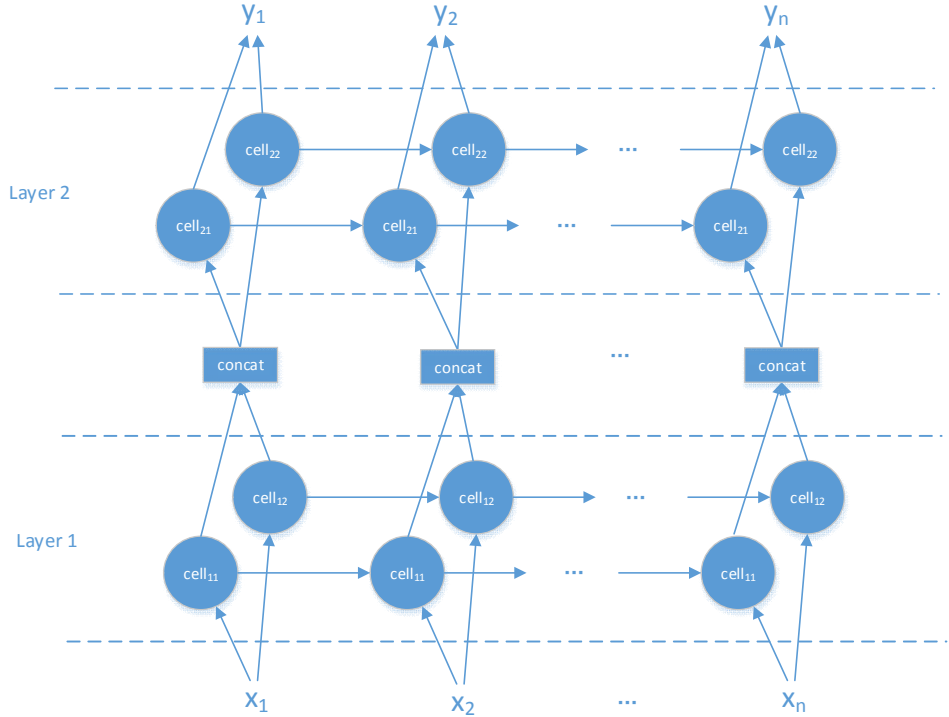

Figure 2. A unrolled 2 layer RNN with 2 parallel cells. $x_i, y_i$ denote the input vector and output vector at time step *i*. Cell*i* denotes the parallel cell *i*.

We evaluate RNN-PCs on two tasks: language modeling and Chinese-English translation. The language modeling task is conducted on Penn Tree Bank (PTB). Our model outperforms state of art systems by decreasing perplexity from 78.6 to 75.3. On the task of Chinese-English machine translation, the proposed model achieves BLEU score X, outperforms the strong baseline model of BLEU score Y.

The rest of the paper is organized as follows. Section 2 describes models, including background models and parallel cells. Section 3 reports experiments. In section 4, based on the task of language modeling, we make some empirical study to observe the behavior of RNN-PCs. Finally, section 5 draws conclusions.

## 2. Methods

### 2.1 Background Models

We briefly go through background models used in the paper, including recurrent neural network(RNN), long-short term memories(LSTM) and their variants.

**Naïve Recurrent Neural Network**

RNN cell is the basic computation component in RNN. A cell is mainly composed by three parts, input vector，hidden states and recurrent connection. Generally, the output of the cell are the hidden states. RNN dynamics can be viewed as a deterministic transition from the past hidden states to current states given current input. Recurrent connection defines how to the transition procedure happens.

We let subscript denote time step and subscript denote layers. Let $h_t^l \in \mathbb{R}^n$ be hidden states of RNN cell in time step t at layer l. Let $T_{n,m}: \mathbb{R}^n \to \mathbb{R}^m$ be a linear transform with a bias, e.g. $y = Wx + b$, for some W and b, where $x \in \mathbb{R}^n$ and $y \in \mathbb{R}^m$. Let $f(.)$ be sigmoid, tanh, relu or other non-linear activate functions. Let $\odot$ be element-wise multiplication.

At time step t in layer l, the hidden state $h_t^l$ is determined by previous hidden state $h_{t-1}^l$ and current input $x_t^l$.

$$h_t^l = f(T_{n,n} h_{t-1}^l + x_t^l)$$

**Long Short-Term Memories**

Long short-term memory (LSTM) (Hochreiter and Schmidhuber 1997) addresses the problem of learning long range dependencies by augmenting the RNN with a memory cell vector $c_t^l \in \mathbb{R}^n$. In this paper, we follow LSTM architecture by Graves (2013). LSTM uses forget gate *f*, input gate *i*, output gate *o* to control the data flow. The equations are as follows.

$$\begin{pmatrix} i \\ f \\ o \\ g \end{pmatrix} = \begin{pmatrix} \text{sigm} \\ \text{sigm} \\ \text{sigm} \\ \text{tanh} \end{pmatrix} T_{2n,4n} \begin{pmatrix} x_t^l \\ h_{t-1}^l \end{pmatrix}$$

$$c_t^l = f \odot c_{t-1}^l + i \odot g$$
$$h_t^l = o \odot \tanh(c_t^l)$$

**RNN Variants**

Two commonly used variants of basic RNN architecture are the Bidirectional RNN and Multilayer RNN.

Bidirectional RNN contains of two RNN cells that are in parallel: one on the input sequence and the other on the reverse of the input sequence. At each time step, the hidden states of forward RNN and backward RNN are concatenated as output.

In Multilayer RNN, there are multilayer of RNN, each layer contains a single RNN cell. The output of lower layer is fed into its upper layer as input.

**2.2 Parallel cells**

The key idea of PCs is to replace a single large cell with several small parallel cells. Figure 3 shows the a basic RNN cell and its counterpart parallel cells solution. The left side is a basic RNN cell whose hidden states has *m* units. Let *RC* be the cell's recurrent connection. Let *x,h* be the input vector and hidden vector respectively. The right side shows a counterpart parallel cells solution. There are *n* small parallel cells, each with *m/n* unit of hidden states. Thus both the left and the right side have equal total unit of hidden states to retain information from the past. The small cells accept input x and generate outputs *h_1,h_2…h_n* respectively. The final output *h* is concatenated by *h_1,h_2…h_n*.

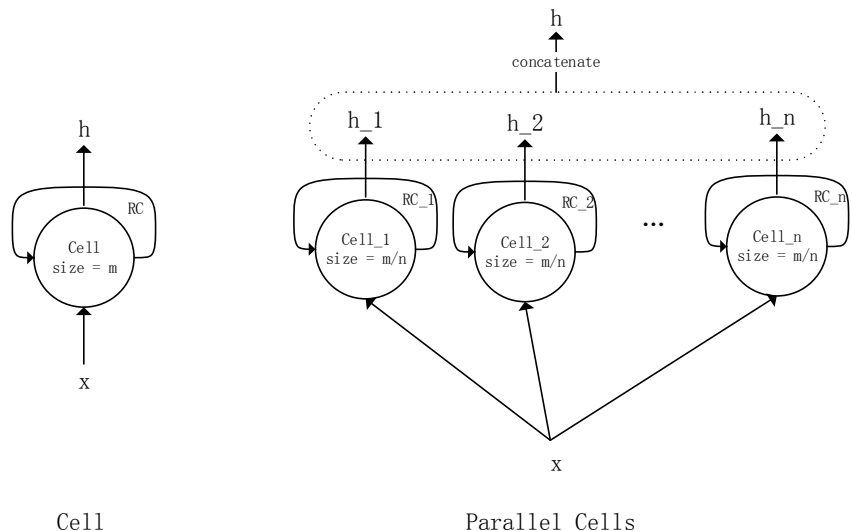

Cell                    Parallel Cells

Figure 3. Rnn cell & counterpart Parallel cells solution.

PCs can be easily used along with LSTM or other complex RNN cells. As shown in the figure, parallel cells do not modify any internal structure of cells. For example, LSTM cell also accepts a input, outputs its hidden units. The mainly modification of LSTM are the memory vector and recurrent connection, which have no impact with PCs.

Compared to naïve RNN, here we add a new hyper-parameter *wide*, denotes the amount of parallel cells. When we set *wide=1*, the model is exactly the same as naïve RNN. Note that we replace one cell with many parallel small cells by equally dividing the units of hidden states. It is really a arbitrary setting just for convenience of turning. Maybe a more delicate splitting method will be more helpful for improving learning ability.

**Parameters amount**

For basic RNN cell with *m* units, the parameters are $T_{m,m}$, with total parameter size $m^2 + m$. Suppose we use PCs with *wide=n*, there are n $T_{\frac{m}{n},\frac{m}{n}}$, with total number of parameters $\frac{m^2}{n} + m$. When m is large, PCs with *wide=n* can reduce the parameter size of the origin RNN cell to about $\frac{1}{n}$.

For LSTM with *m* units, the parameters are $T_{2m,4m}$ for calculating *i,f,o,g*, with total parameter size $8m^2 + 4m$. Suppose we use parallel cells with *wide=n*, there are n $T_{\frac{4m}{n},\frac{2m}{n}}$, with total number of parameters $\frac{8m^2}{n} + 4m$. When m is large, Pcs with *wide=n* can reduce the parameters size of the origin LSTM cell to about $\frac{1}{n}$.

**3.Experiments**

We present results in language modeling, machine translation and part-of-speech(POS) tagging.

**3.1 language modeling**

**Task and Dataset**

We conduct word-level word prediction experiments on the Penn Tree Bank (PTB) dataset. The data is from Tomas Mikolov's webpage[1]. We use exactly the same training data and test data as other researchers(Zaremba et al., 2014; Kim et al., 2016), about 90% for training and 10% for testing. The vocabulary size is 10k. We use a *UNK* symbol for the rest of rare words.

**Model and Training Details**

Zaremba et al.(2014) achieved state of art performance on PTB with a two layer regularized LSTM. At each time step, word is mapped to a fix length word vector and then fed into LSTM. LSTM outputs a probability distribution of the next word. Our models are regularized LSTM with PCs.

For convenience of comparison, we follow most of Zaremba's settings. LSTMs in all experiments have two layers and are unrolled in 35 steps. The initial states are 0 and we use the final states of previous batch as the initial states of the next batch. The size of batch is 20. We use cross entropy loss function and stochastic gradient decent for optimization. Model parameters are initialized uniformly between [-0.04,0.04]. We train the model with learning rate 1 in first 14 epochs, and apply weight decay by 1/1.15 in the rest 41 epochs. We clip the norm of the gradients at 5(Mikolov et al., 2010). We dropout 65% of recurrent connections(Srivastava et al.,

---
[1] http://www.fit.vutbr.cz/~imikolov/rnnlm/simple-examples.tgz

2014) to avoid overfitting.

**Results**

Table 1 shows results of previous systems and our model. Zaremba et al.(2014) achieved perplexity 78.4, the state of art result for language modeling on PTB, and significantly outperformed all previous works. Their model has 1500 hidden units. Kim et al.(2016) proposed a model with pure character input, and got comparable perplexity 78.9. Our model achieves perplexity 75.3 when *wide=3*, hidden units are 1950, decreases 3.1 perplexity than state of art results.

| Model | Test Perplexity |
|---|---|
| LSTM-1500h(Zaremba et al., 2014) | 78.4 |
| LSTM-Char(Kim et al, 2016) | 78.9 |
| **Parallel_Cells_LSTM_1950h_3wide** | **75.3** |

Table 1: Perplexity of our model versus previous works on PTB

In table 2, We show the performance of our model with different *wide*. When *wide=1*, the model is a plain 2 layer LSTM. When *wide=3* to *10*, the models get similar performance. Note that when *wide=10*, the LSTM cell contains about only 10% of parameters, while reduce 4.5 perplexity than plain LSTM. However, continue to add parallel cells will harm the performance. For *wide=15*, the perplexity increases to 77.5. We suggest that use PCs will reduce the unnecessary connections from unrelated features. But when the cell is getting too small to afford a necessary feature sets, the overall performance will decrease.

For plain LSTM, setting the units of hidden states larger than 1500 will not improve the performance. From 1500 to 1950, the perplexity grows from 78.4 to 80.0. However, parallel cells can still benefit from increasing units of hidden states.

| wide | Recurrent Parameters Size | perplexity |
|---|---|---|
| 1 | 30.4m | 80.0 |
| 2 | 15.2m | 76.4 |
| 3 | 10.1m | 75.3 |
| 4 | 7.6m | 75.4 |
| 5 | 6.1m | 75.6 |
| 10 | 3.4m | 75.5 |
| 15 | 2.4m | 77.5 |

Table 2: the perplexity on different *wide*. The models are all 1950h and 2 lays.

### 3.2 Machine Translation

**Task and Dataset**

We conduct experiments on a Chinese-English translation task. The purpose of this experiment is to evaluate that whether PCs has a substantial improvement on large RNN. For consideration of speed, we only include training and testing samples when both sides contain less than 50 words. We limit the source and target vocabularies to the most frequent 30K words in Chinese and English, including a UNK symbol for other words, a PAD symbol for padding and a EOS for end of the sentence. Our training data for the translation task consists of 1.25M sentence pairs extracted from

LDC corpora[2]. The development dataset is NIST 2002 dataset, contains 824 sentence pairs. The test datasets are the NIST 2003, 2004, 2005 and 2006 datasets, contains 4910 sentence pairs. We use the case insensitive 4-gram NIST BLEU score (Papineni et al., 2002) for evaluating the translation task.

**Model and Training Details**

Bahdanau et al.(2015) proposed a encoder-decoder model with attention for machine translation. Both the encoder and decoder were RNNs and the cell type was Gate Recurrent Units(GRU). The encoder RNN read source words one by one, and the final hidden states encoded all the information of the source sequences. The hidden states were then fed into decoder RNN. At each time step, took states of aligned source states and previous generated target words as inputs, the decoder output current target word. They used bidirectional RNN for the all layers of the encoder to catch both forward and backward information. Wu et al.(2016) used LSTM cells for translation. For speed and parallel running consideration, they only used bidirectional RNN in the first layer of the encoder. Their systems were also comparable to other state of art systems. The baseline model for machine translation in this paper is the same as Bahdanau' work, while there are two minus differences. First, only the first layer of encoder is bidirectional. Second, the cells used in RNNs are LSTM.

We implement PCs for both encoder and the decoder RNN. The bidirectional layer and other layers in encoder does not share parameters. The encoder contains 4 layers, 1 bidirectional bottom layer and 3 single directional layers. The decoder contains 3 layers. The word embedding dimension and size of hidden states are 1024. We use cross entropy loss function and stochastic gradient decent for optimization. The batch size is 32. We train the model with initial learning rate 0.5, and apply weight decay by 0.93 in every 3000 batches if the cost of validation set does not decrease. We clip the norm of the gradients at 5. We dropout 20% of recurrent connections. We train each model for about 800k batches.

**Results**

Table 3 shows the results of baseline model and our model. For the baseline model, it takes about 824000 batches to get the best bleu score in dev set. For translation model with parallel cells, it takes about 714000 batches to get best bleu score. With parallel cells, the translation model gets a improvement of 0.39 on bleu score.

| Model | NIST2002(DEV) | NIST2003 | NIST2004 | NIST2005 | overall |
|---|---|---|---|---|---|
| Bi-LSTM + Attention(Baseline) | 28.08 | 25.71 | 29.81 | 25.90 | 27.14 |
| RNN_PCs + Bi-LSTM + Attention(Wide =3) | 28.39(+0.31) | 26.04(+0.33) | 30.26(+0.45) | 26.28(+0.38) | 27.53(+0.39) |

Table 3: The results of Machine Translation

As 2 examples, table 4 shows some translation samples in the test sets translated by baseline model *vs* by parallel cells model.

| 1 | Chinese | 印度 国防部 在 其 发表 的 一 项 远景 规划 中 也 认 为 , 中国 对 印度 没有 军事 威胁 。 |
|---|---|---|
| | Golden Translation | india 's ministry of defense also said in its long-term plan that china has not posed any military threat to india . |
| | Baseline Translation | in his recent planning plan published by india defense |

---
[2] The corpora include LDC2002E18, LDC2003E07, LDC2003E14, Hansards portion of LDC2004T07, LDC2004T08 and LDC2005T06.

| | | department , india also held that china did not pose any military threat to india . |
|---|---|---|
| | Parallel cells(wide=3) | a long-term plan issued by indian defense ministry also believes that china has no military threat to india . |
| 2 | Chinese | 蒙方 感谢 中方 多年 来 对 蒙古 提供 的 援助 。 |
| | Golden Translation | the mongolian side thanked china for its assistance to mongolia over the years . |
| | Baseline Translation | the mongolian side thanked china for its many years to aid the mongolian government . |
| | Parallel cells(wide=3) | the mongolian side thanked china for its assistance to mongolia over the past years . |

Table 4: Example of Translation Results

## 4 Empirical Study of PCs

Previous, we have shown PCs can substantially improve the performance of RNN with less parameters. In this section, We make a empirical study on language modeling task to show whether parallel cells work in the manner we expected, e.g. difference cells learning different set of features. Also, we compare the performance of PCs with model average techniques.

### 4.1 Different cells for Different Set of Features

Unlike conditional random fields or support vector machine, neural network does not have explicit features. Moreover, texts are not as easy to be visualized as pictures to see what kind of features have been extracted. As a result, we investigate in a indirect way by seeing how result changes when we mask different cells.

We suppose that predict a word requires a set of features offered by cells. As a result, when we mask a cell, if the prediction of target word depends heavily on the features in the cell, the perplexity will increase sharply, and vice versus.

We run a single layer LSTM, with 420 unit of hidden states and *wide=3*, dropout rate=0.2 on PTB. On test set, we mask the parallel cells in turns to see what changes. Figure 4 shows how we mask the cells.

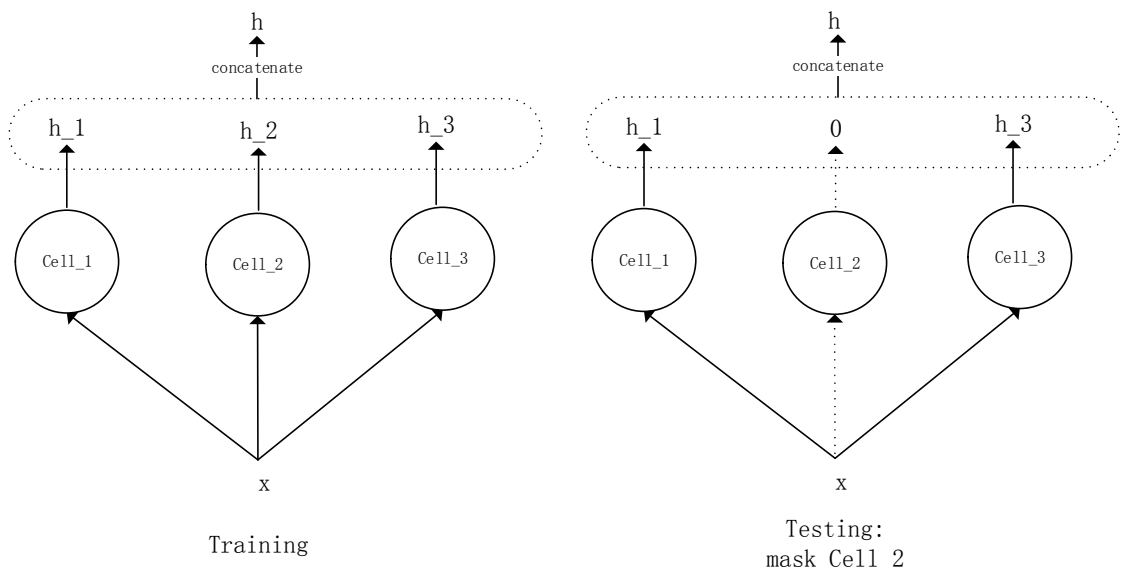

Figure 4: Mask the cell. The training procedure makes no difference as previous RNN-PCs. On testing, the right

side of the figure shows that we mask cell_2 by setting its hidden states to 0.

Figure 5 show three group of samples selected from test set. The words are all selected from about top 100 most frequency word set. It can be seen that for similar words, masked models have similar performance on perplexity. For example, when we consider copular verbs, generally, mask1 gets the lowest average perplexity on all words, while mask2 gets almost the highest perplexity. We can conclude that when predicting copular verbs, cell1 provides less important features than cell2.

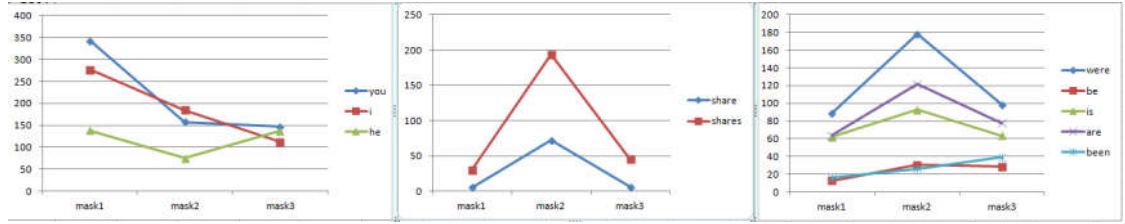

Figure 5: Results of masked models. The vertical axis shows the average perplexity in test set. The horizon axis mask*i* represents the model masking the cell i.

### 4.2 comparison with model averaging

A PCs-RNN with *m* hidden units and *n* parallel cells consumes as much computing and memory resource as *n* small naïve RNN with m/n hidden units. Thus, we want to compare the performance between PCs-RNN and its counterpart ensemble models that consumes the same resource.

Table 3 compares the results of model averaging. 2 model1 cost as much resource as 6 model3. However, 2 model1 outperform 10 model3 on model averaging. 38 model2 achieve previous state of art perplexity on model averaging. However, we outperform that by using only 10 model1.

We can conclude that, after model averaging, PCs can still substantially outperforms naïve RNN with equivalent, or less computing and memory resource.

|  | **Model** | **Test Perplexity** |
| --- | --- | --- |
| Single Model | Model1 | **75.3** |
|  | Model2 | 82.7 |
|  | Model3 | 78.4 |
| Model Averaging | 10 model3 | 72.0 |
|  | 2 model1 | **71.7** |
|  | 10 model2 | 69.5 |
|  | 38 model2 | 68.7 |
|  | 10 model1 | **68.6** |

Table 3: Results of Model Averaging.

Model1: *Parallel_Cells_LSTM_1950h_3wide* in table 1, with 3 parallel cells and 1950 hidden units.

Model2: *LSTM-1500h* in table 1, the naïve LSTM with 1500 hidden units.

Model3: A naïve LSTM with 650 hidden states.

All data about model2 and model3 are the results reported in Zaremba et al.(2014).

### 5 Conclusion

In this paper, we proposed Parallel Cells technique for RNN. Parallel cells can reduce the impact from unrelated features coming from recurrent connections. On all 3 evaluation tasks, models with PCs get significantly improvement on results with less parameters. We investigated the results of language modeling task by masking cells. The results show a strong evidence that different cells are coping with different set of features. Also, we found that even after model averaging, PCs can still beat the baseline model, and get new state of art perplexity.

There are much work to be done in the line of PCs. For example, parallel cells with different hidden units maybe better, because a equally dividing strategy is really arbitrary. Also, we can test the performance of PCs on many downstream models based on RNN.

References

[Mikolov et al., 2010] T. Mikolov, M. Karafiát, L. Burget, J. Cernocký, and S. Khudanpur. Recurrent neural network based language model. In INTERSPEECH, pages 1045–1048, 2010.

[Zaremba et al., 2014] W. Zaremba, I. Sutskever, and O. Vinyals. Recurrent neural network regularization. In arXiv:1409.2329, 2014.

[Luong and Manning, 2016] Luong M T, Manning C D, Minh-Thang Luong and Christopher D. Manning. 2016. Achieving open vocabulary neural machine translation with hybrid word-character models. In ACL, pages 1054-1063, 2016

[Wu et al., 2016] Yonghui Wu, Mike Schuster, Zhifeng Chen, Quoc V Le, Mohammad Norouzi, Wolfgang Macherey, Maxim Krikun, Yuan Cao, Qin Gao, Klaus Macherey, et al. 2016. Google's neural machine translation system: Bridging the gap between human and machine translation. In arXiv:1609.08144.

[Hochreiter and Schmidhuber, 1997] Hochreiter, Sepp and Schmidhuber, Jürgen. Long short-term memory. Neural computation, 9(8):1735–1780, 1997.

[Cho et al., 2014]Kyunghyun Cho, Bart van Merrienboer, Dzmitry Bahdanau, Yoshua Bengio. On the Properties of Neural Machine Translation: Encoder-Decoder Approaches. In EMNLP 2014, pages 103-111.

[Lecun et al., 1998] Y. LeCun, L. Bottou, Y. Bengio, and P. Haffner. Gradient-based learning applied to document recognition. Proceedings of the IEEE, 86(11):2278–2324, 1998.

[Burt, 1988] Burt P J. Attention mechanisms for vision in a dynamic world. International Conference on Pattern Recognition. Pages 977-987, 1988, vol.2.

[Denil et al., 2012] Denil M, Bazzani L, Larochelle H, et al. Learning where to attend with deep architectures for image tracking. Neural Computation, pages 2151-2184, 2012, 24(8).

[Bahdanau et al., 2015] Dzmitry Bahdanau, Kyunghyun Cho, and Yoshua Bengio. 2015. Neural machine translation by jointly learning to align and translate. ICLR 2015.

[Luong et al., 2015] Thang Luong, Hieu Pham, and Christopher D. Manning. Effective Approaches to Attention based Neural Machine Translation. In Proceedings of the 2015 Conference on Empirical Methods in Natural Language Processing, pages 1412–1421.

[Chung et al., 2014] Chung J, Gulcehre C, Cho K H, et al. Empirical Evaluation of Gated Recurrent Neural Networks on Sequence Modeling. Eprint Arxiv, 2014.

[Graves, 2013] Graves, Alex. Generating sequences with recurrent neural networks. arXiv preprint arXiv:1308.0850, 2013.

[Kim et al., 2016]Yoon Kim, Yacine Jernite, David Sontag and Alexander M. Rush. Character-aware neural language models. In AAAI 2016, pages 2741-2749.

[Srivastava et al., 2014]Srivastava N, Hinton G, Krizhevsky A, et al. Dropout: a simple way to prevent neural networks from overfitting[J]. Journal of Machine Learning Research, 2014, 15(1):1929-1958.

[Papineni et al., 2002] Kishore Papineni, Salim Roukos, Todd Ward, and Wei-Jing Zhu. BLEU: a method for automatic evaluation of machine translation. In ACL 2002, 311-318.

[Hinton et al., 2012] G. E. Hinton, N. Srivastava, A. Krizhevsky, I. Sutskever, and R. R. Salakhutdinov. Improving neural networks by preventing coadaptation of feature detectors. arXiv:1207.0580, 2012.

[Collins,2002]Michael Collins. 2002. Discriminative Training Methods for Hidden Markov Models: Theory and Experiments with Perceptron Algorithms. In Proceedings of EMNLP, pages 1–8, Stroudsburg, PA, USA.

[Wang et al., 2015] Wang P, Qian Y, Soong F K, et al. A Unified Tagging Solution: Bidirectional LSTM Recurrent Neural Network with Word Embedding. arXiv preprint arXiv:1511.00215, 2015.